\documentclass{article}

\PassOptionsToPackage{numbers, sort&compress}{natbib}
\usepackage[preprint]{neurips_2025}

\usepackage{amsmath}
\usepackage{amssymb}

\usepackage{algorithm}
\usepackage{algorithmic}

\usepackage{booktabs}
\usepackage{siunitx}
\DeclareSIUnit[number-unit-product={}]{\x}{\ensuremath{\times}}

\usepackage{graphicx}
\usepackage{tikz}
\usetikzlibrary{arrows.meta,calc,decorations.markings}

\usepackage{microtype}
\usepackage{enumitem}
\setlist{leftmargin=*}

\usepackage{hyperref}
\usepackage{url}

\title{ScribeTokens: Fixed-Vocabulary Tokenization of Digital Ink}

\author{Douglass Wang \\
  Independent Researcher \\
  \texttt{douglasswng@gmail.com} \\
  \url{https://github.com/douglasswng/scribe-tokens}
}

\begin{document}

\maketitle

\begin{abstract}
  Digital ink---the coordinate stream captured from stylus or touch input---lacks a unified representation. Continuous vector representations produce long sequences and suffer from training instability, while existing token representations require large vocabularies, face out-of-vocabulary issues, and underperform vectors on recognition. We propose ScribeTokens, a tokenization that decomposes pen movement into unit pixel steps. Together with two pen-state tokens, this fixed 10-token base vocabulary suffices to represent any digital ink and enables aggressive BPE compression. On handwritten text generation, ScribeTokens dramatically outperforms vectors (17.33\% vs.\ 70.29\% CER), showing tokens are far more effective for generation. On recognition, ScribeTokens is the only token representation to outperform vectors without pretraining. We further introduce next-ink-token prediction as a self-supervised pretraining strategy, which consistently improves recognition across all token-based models and accelerates convergence by up to \SI{83}{\x}. With pretraining, ScribeTokens achieves the best recognition results across all representations on both datasets (8.27\% CER on IAM, 9.83\% on DeepWriting).
\end{abstract}

\section{Introduction}
\label{sec:introduction}

\emph{Digital ink}---the coordinate stream captured from stylus or touch input---is a structured sequential modality underlying applications from handwriting recognition~\citep{graves2009novel} and mathematical expression parsing~\citep{chan2000mathematical} to sketch synthesis~\citep{ha2018neural} and handwriting generation~\citep{graves2013generating}. Unlike offline handwriting images, digital ink preserves temporal writing dynamics as sequences of strokes, each comprising a sequence of $xy$-coordinates. How this two-level structure is flattened into a single sequence for modeling directly affects both sequence length and training stability---and, in turn, inference speed and task performance.

Despite growing interest in digital ink modeling, existing representations each carry significant limitations. Vector representations encode ink as sequences of continuous coordinates with binary pen-up/down flags~\citep{graves2009novel, graves2013generating, ha2018neural}. While widely used, they produce long sequences and require careful design choices for input normalization and sequence handling. For generation, they rely on mixture density networks that suffer from training instability and hard-to-interpret likelihood values~\citep{cui2019multimodal,makansi2019overcoming,bishop1994mixture}. Token representations have emerged as an alternative~\citep{fadeeva2024representing, ribeiro2020sketchformer}, enabling compression via Byte-Pair Encoding (BPE)~\citep{sennrich2016neural} and more stable training through cross-entropy loss. However, existing tokenization methods carry their own drawbacks: out-of-vocabulary (OOV) issues, large base vocabularies that scale with canvas resolution, or fragile syntax where malformed outputs do not decode into valid ink. As we show, they also underperform vector representations on recognition benchmarks.

\begin{figure}[t]
  \centerline{\includegraphics[width=\textwidth]{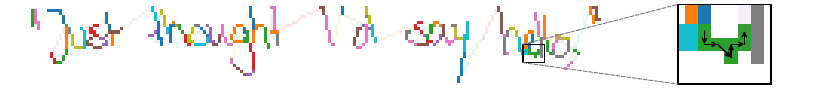}}
  \caption{ScribeTokens representation of a handwritten sentence. Pen strokes are decomposed into unit directional steps via Bresenham's algorithm, then compressed with BPE. Each color denotes a distinct BPE token; faint colors indicate pen-in-air movement between strokes. The zoom shows the sequence of arrows making up an example token.}
  \label{fig:scribe}
\end{figure}

We propose \emph{ScribeTokens}, a token representation with no OOV by construction, a fixed base vocabulary of just 10 tokens, and robust syntax: every token sequence decodes into valid ink. Our key insight is to decompose pen movements into unit steps between adjacent pixels via Bresenham's line algorithm~\citep{bresenham1965algorithm}. Each step is encoded as one of eight direction tokens inspired by Freeman chain codes~\citep{freeman1961encoding}, and two additional tokens signal pen-up and pen-down. Every token carries an intuitive physical interpretation. Because ScribeTokens encodes paths rather than points, it yields a canonical representation invariant to sampling rate or point density. BPE applied over this base vocabulary yields high compression while preserving the OOV-free property, since any unseen pattern can always be decomposed into base tokens (Figure~\ref{fig:scribe}).

We evaluate ScribeTokens on handwritten text recognition and generation across two datasets~\citep{liwicki2005iam, aksan2018deepwriting}. Our experiments reveal three key findings. First, token representations are far more effective than vectors for generation---ScribeTokens achieves 17.33\% character error rate (CER) compared to 70.29\% for vectors on sentence-level data. Second, next-ink-token prediction~\citep{radford2018improving, brown2020language} is an effective self-supervised pretraining strategy for tokens, consistently improving recognition and accelerating convergence by up to \SI{83}{\x}; for vectors, pretraining can even deteriorate performance. Third, ScribeTokens with pretraining achieves the best recognition results on both datasets (8.27\% and 9.83\% CER).

\paragraph{Contributions.} (1) We propose ScribeTokens, a canonical, OOV-free tokenization of digital ink with a fixed 10-token base vocabulary and robust syntax, enabling aggressive BPE compression.\footnote{An accompanying Hugging Face~\citep{wolf2020huggingface} library for fast tokenization and tokenizer training is available at \url{https://github.com/douglasswng/tokink}.} ScribeTokens is the only token representation to outperform vectors on recognition without pretraining.
(2) We show that next-ink-token prediction is an effective self-supervised pretraining strategy for tokens, consistently improving recognition and yielding up to \SI{83}{\x} faster convergence. ScribeTokens with pretraining achieves the best recognition results on both datasets and the best generation results in the data-limited setting.

\section{Related Work}
\label{sec:related}

\begin{table}[t]
  \caption{Digital ink representations for a two-stroke example. Colors distinguish the {\color{blue}first stroke}, {\color{gray}pen-in-air}, and {\color{red}second stroke}. \emph{Point-3} encodes offsets with a binary pen flag; \emph{Point-5} extends this with a one-hot pen state that additionally signals end of sequence; \emph{AbsTokens} and \emph{RelTokens} discretize absolute and relative coordinates into tokens; \emph{TextTokens} serialize offsets as character sequences; \emph{ScribeTokens} (Ours) decompose strokes into unit directional steps via Bresenham's algorithm.}
  \label{tab:reprs}
  \centering
  \begin{tabular}{cc}
  \toprule
  Representation & Value \\
  \midrule
  {Digital Ink} & $\color{blue}((0, 0), (1, 0))$, $\color{red}((2, 1), (4, -1))$ \\
  \addlinespace
  Visualization &
  \begin{tikzpicture}[scale=0.618, baseline]
    \draw[gray!30, thin] (-0.5,-1.5) grid (4.5,1.5);

    \draw[->, gray!80, thin] (-0.5,0) -- (4.7,0) node[right, gray!80] {$x$};
    \draw[->, gray!80, thin] (0,-1.5) -- (0,1.5) node[above, gray!80] {$y$};

    \foreach \x in {0,1,2,3,4}
    \draw (\x,0) node[below, font=\small] {$\x$};

    \foreach \y in {-1,0,1}
    \draw (0,\y) node[left, font=\small] {$\y$};

    \filldraw[blue] (0,0) circle (2pt);
    \filldraw[blue] (1,0) circle (2pt);
    \filldraw[red] (2,1) circle (2pt);
    \filldraw[red] (4,-1) circle (2pt);

    \draw[blue, thick, decoration={markings, mark=at position 0.5 with {\arrow{>}}}, postaction={decorate}] (0,0) -- (1,0);
    \draw[gray, thick, dashed, decoration={markings, mark=at position 0.5 with {\arrow{>}}}, postaction={decorate}] (1,0) -- (2,1);
    \draw[red, thick, decoration={markings, mark=at position 0.5 with {\arrow{>}}}, postaction={decorate}] (2,1) -- (4,-1);
  \end{tikzpicture}\\
  \addlinespace
  {Point-3} & $\color{blue}(1, 0, 1)$, $\color{gray}(1, 1, 0)$, $\color{red}(2, -2, 1)$\\
  \addlinespace
  Point-5 & $\color{blue}(1, 0, 1, 0, 0)$, $\color{gray}(1, 1, 0, 1, 0)$, $\color{red}(2, -2, 0, 0, 1)$\\
  \addlinespace
  AbsTokens & \texttt{\color{blue}[(0, 0)]}, \texttt{\color{blue}[(1, 0)]}, \texttt{\color{blue}[UP]}, \texttt{\color{red}[(2, 1)]}, \texttt{\color{red}[(4, -1)]}, \texttt{\color{red}[UP]}\\
  \addlinespace
  RelTokens & \texttt{\color{blue}[(1, 0)]}, \texttt{\color{blue}[UP]}, \texttt{\color{gray}[(1, 1)]}, \texttt{\color{red}[(2, -2)]}, \texttt{\color{red}[UP]}\\
  \addlinespace
  TextTokens & \texttt{\color{blue}[1]}, \texttt{\color{blue}[\textvisiblespace]}, \texttt{\color{blue}[0]}, \texttt{\color{blue}[UP]}, \texttt{\color{gray}[1]}, \texttt{\color{gray}[\textvisiblespace]}, \texttt{\color{gray}[1]}, \texttt{\color{gray}[\textvisiblespace]}, \texttt{\color{red}[2]}, \texttt{\color{red}[\textvisiblespace]}, \texttt{\color{red}[-]}, \texttt{\color{red}[2]}, \texttt{\color{red}[UP]}\\
  \midrule
  ScribeTokens (Ours) & \texttt{\color{blue}[DOWN]}, \texttt{\color{blue}[$\rightarrow$]}, \texttt{\color{blue}[UP]}, \texttt{\color{gray}[$\nearrow$]}, \texttt{\color{red}[DOWN]}, \texttt{\color{red}[$\searrow$]}, \texttt{\color{red}[$\searrow$]}, \texttt{\color{red}[UP]}\\
  \bottomrule
\end{tabular}

\end{table}

\subsection{Vector Representations}
Vector representations model digital ink as a sequence of continuous vectors~\citep{graves2013generating,dai2023disentangling,ha2018neural,carbune2020fast}. For generation, the next $xy$-coordinate is modeled as a mixture of 2-dimensional Gaussians, with discrete events (such as pen-up) modeled by a categorical distribution~\citep{graves2013generating,dai2023disentangling,ha2018neural}.

\emph{Point-3}~\citep{graves2013generating} encodes each point as $(\Delta x, \Delta y, p)$, where $\Delta x, \Delta y$ are offset coordinates from the previous point and $p$ is a binary pen-up indicator. \emph{Point-5}~\citep{ha2018neural} extends this to $(\Delta x, \Delta y, p_1, p_2, p_3)$, using a one-hot pen state across three mutually exclusive states: pen-down, pen-up (end of stroke), and end of sequence. Table~\ref{tab:reprs} illustrates each representation for a two-stroke example.

Vector representations, while natural, suffer from several drawbacks: (1) the lack of compression often necessitates truncating long sequences~\citep{graves2013generating},\footnote{While \citet{carbune2020fast} present a vector representation achieving significant compression, their approach is not applicable to generation tasks.} (2) many design choices must be made for input normalization~\citep{graves2013generating, aksan2018deepwriting, dai2023disentangling, zhang2017drawing} and sequence initiation/termination~\citep{graves2013generating, dai2023disentangling}, and (3) mixture density networks used for generation require additional hyperparameters such as mixture count and loss weighting, suffer from numerical instability~\citep{cui2019multimodal} and mode collapse~\citep{makansi2019overcoming}, and produce negative log-likelihood values~\citep{bishop1994mixture} that are difficult to interpret.

\subsection{Token Representations}
Token representations treat digital ink as a sequence of discrete tokens, which addresses the challenges faced by vector representations: (1) merging algorithms such as BPE effectively reduce sequence lengths, (2) \texttt{[START]} and \texttt{[END]} tokens eliminate design choices for sequence initiation and termination, and (3) cross-entropy loss does not introduce additional hyperparameters and allows for stable training with interpretable loss values~\citep{goodfellow2016deep}.

\emph{AbsTokens}~\citep{ribeiro2020sketchformer} treats each pixel coordinate as a token, with \texttt{[UP]} for pen-up. \emph{RelTokens}~\citep{ribeiro2020sketchformer} is similar but uses relative offsets $(\Delta x, \Delta y)$ as tokens. \emph{TextTokens}~\citep{fadeeva2024representing} serializes each offset as its decimal string (digits, minus signs, and spaces).

Existing token representations, however, have notable drawbacks. AbsTokens and RelTokens suffer from out-of-vocabulary (OOV) issues, mapping unseen coordinates to \texttt{[UNKNOWN]}, and require large base vocabularies that scale with canvas resolution. TextTokens avoids both problems but produces sequences with a fragile syntax: an autoregressive model can emit malformed sequences that do not parse into valid coordinate pairs. ScribeTokens sidesteps all three issues: no OOV by construction, a fixed 10-token base vocabulary, and a robust syntax in which every token sequence is valid.

\subsection{Pretraining}
In natural language processing, self-supervised pretraining on large unlabeled corpora followed by supervised fine-tuning has become the standard paradigm, consistently outperforming training on labeled data alone~\citep{radford2018improving, devlin2019bert, brown2020language, raffel2020exploring}. This paradigm has also been extended to speech~\citep{du2024cosyvoice} and music~\citep{dhariwal2020jukebox}. In digital ink, pretraining has been explored primarily for sketches: \citet{lin2020sketch} apply BERT-style masked prediction over point positions and pen states, \citet{bhunia2021vectorization} propose cross-modal learning between raster and vector representations, and \citet{tiwari2024sketchgpt} use autoregressive next-token prediction on discrete sketch primitives.

However, these methods target sketches rather than handwriting, and each adopts a single ink representation without examining how this choice affects pretraining. Self-supervised pretraining for online handwriting remains largely unexplored. We address this gap by investigating next-ink-token prediction as a self-supervised pretraining strategy for online handwriting across multiple ink representations. Notably, we find that pretraining efficacy depends strongly on the choice of representation.

\section{Methods}
\label{sec:methods}

\subsection{Preprocessing}

We define digital ink $\mathcal{I} = (S_j)_{j=1}^M$ as a sequence of strokes, where each stroke $S_j = (p_i^{(j)})_{i=1}^{n_j}$ is a sequence of continuous $xy$-coordinates. Since token representations are intrinsically discrete, we quantize the coordinates to a grid. More aggressive quantization reduces token sparsity and improves BPE compression, at the cost of increased staircase artifacts in the discretized ink.

Following~\citet{ribeiro2020sketchformer}, we round each coordinate to the nearest point on a uniform grid with spacing $\delta > 0$:
\begin{equation*}
  (x_i, y_i) \mapsto \left(\mathrm{round}\left(\frac{x_i}{\delta}\right),\; \mathrm{round}\left(\frac{y_i}{\delta}\right)\right).
\end{equation*}
This produces an \emph{integer ink} on which all subsequent methods operate.

\subsection{Bresenham Decomposition}

The core idea behind ScribeTokens is to represent pen strokes as sequences of unit directional steps on a discrete grid. This is achieved by combining two classical algorithms: Freeman chain codes~\citep{freeman1961encoding} for directional encoding and Bresenham's line algorithm~\citep{bresenham1965algorithm} for integer rasterization.

A Freeman chain code encodes a path on a discrete grid as a sequence of unit steps in eight directions---the four cardinal ($\rightarrow$, $\uparrow$, $\leftarrow$, $\downarrow$) and four diagonal ($\nearrow$, $\nwarrow$, $\swarrow$, $\searrow$)---making it a lossless encoding for any path between adjacent pixels. However, consecutive points in ink are generally not adjacent. To bridge non-adjacent points, we rasterize the straight-line segment between them using Bresenham's line algorithm, which computes the optimal sequence of adjacent grid cells.

\begin{figure}[t]
  \centering
  \begin{tikzpicture}[scale=0.45]
  \draw[gray!40] (0,0) grid (13,7);

  \coordinate (start) at (1.5,5.5);
  \coordinate (end) at (11.5,1.5);

  \foreach \x/\y in {1/5, 2/5, 3/4, 4/4, 5/3, 6/3, 7/3, 8/2, 9/2, 10/1, 11/1} {
    \fill[blue!15] (\x,\y) rectangle (\x+1,\y+1);
  }

  \draw[gray!40] (0,0) grid (13,7);

  \fill[green!50!black] (start) circle (0.15);
  \fill[red!70!black] (end) circle (0.15);

  \coordinate (prev) at (start);
  \foreach \dx/\dy/\sym in {1/0/\rightarrow, 1/-1/\searrow, 1/0/\rightarrow, 1/-1/\searrow, 1/0/\rightarrow, 1/0/\rightarrow, 1/-1/\searrow, 1/0/\rightarrow, 1/-1/\searrow, 1/0/\rightarrow} {
    \coordinate (next) at ([shift={(\dx,\dy)}]prev);
    \node[inner sep=0pt, yshift=-0.15ex] at ($(prev)!0.5!(next)$) {$\sym$};
    \coordinate (prev) at (next);
  }

  \node[anchor=north] at (6.5,-0.5) {%
    \footnotesize$\mathrm{BD}\bigl(\tikz[baseline=-0.5ex, scale=0.45]{\fill[green!50!black] (0,0) circle (0.15);},\;%
    \tikz[baseline=-0.5ex, scale=0.45]{\fill[red!70!black] (0,0) circle (0.15);}\bigr)%
  \;=\;\rightarrow\;\searrow\;\rightarrow\;\searrow\;\rightarrow\;\rightarrow\;\searrow\;\rightarrow\;\searrow\;\rightarrow$};
\end{tikzpicture}
  \caption{Bresenham Decomposition of a line segment between two grid points (start~\protect\tikz[baseline=-0.5ex, scale=0.45]{\protect\fill[green!50!black] (0,0) circle (0.15);}, end~\protect\tikz[baseline=-0.5ex, scale=0.45]{\protect\fill[red!70!black] (0,0) circle (0.15);}). The segment is rasterized into adjacent grid cells via Bresenham's algorithm, then encoded as a sequence of Freeman chain code directions.}
  \label{fig:bresenham}
\end{figure}
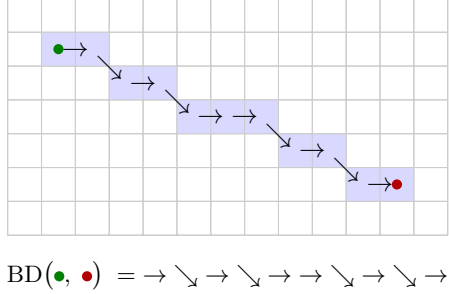

Given two integer points---possibly non-adjacent---we first rasterize the segment between them using Bresenham's algorithm, then encode transitions between consecutive rasterized pixels as Freeman chain code directions. We call this composition \emph{Bresenham Decomposition} ($\mathrm{BD}$). The result is a deterministic sequence of direction tokens uniquely determined by the two endpoints. Figure~\ref{fig:bresenham} illustrates this process.

\subsection{ScribeTokens}
\label{sec:scribetokens}

To tokenize an integer ink $\mathcal{I} = (S_j)_{j=1}^M$, we combine the eight direction tokens with two pen-state tokens: \texttt{[DOWN]} (pen-down) and \texttt{[UP]} (pen-up), forming a fixed vocabulary of 10 base tokens. Each stroke is delimited by \texttt{[DOWN]} and \texttt{[UP]}, and consecutive points---both within strokes and during pen-in-air transitions---are encoded via Bresenham Decomposition. Algorithm~\ref{alg:scribetokens} details the full procedure.

\begin{algorithm}[t]
  \caption{ScribeTokens tokenization}
  \label{alg:scribetokens}
  \begin{algorithmic}[1]
  \REQUIRE Integer ink $\mathcal{I}=(S_j)_{j=1}^M$, each $S_j = (p_i^{(j)})_{i=1}^{n_j}$ a sequence of integer coordinates.
  \ENSURE Token sequence $T = (t_i)_{i=1}^n$
  \STATE $T \leftarrow ()$
  \FOR{$j = 1$ to $M$}
  \STATE \textit{// Begin stroke $j$ (pen touches surface)}
  \STATE Append \texttt{[DOWN]} to $T$
  \FOR{$i = 1$ to $n_j-1$}
  \STATE Append tokens from $\mathrm{BD}(p_i^{(j)},\; p_{i+1}^{(j)})$ to $T$ \COMMENT{within-stroke movement}
  \ENDFOR
  \STATE \textit{// End stroke $j$ (pen lifts off surface)}
  \STATE Append \texttt{[UP]} to $T$
  \STATE \textit{// Movement to next stroke (pen in air)}
  \IF{$j < M$}
  \STATE Append tokens from $\mathrm{BD}(p_{n_j}^{(j)},\; p_1^{(j+1)})$ to $T$ \COMMENT{between-stroke movement}
  \ENDIF
  \ENDFOR
  \RETURN $T$
\end{algorithmic}
\end{algorithm}

\paragraph{Sampling invariance.} An important property of ScribeTokens is sampling invariance: digital inks that rasterize identically on a discrete grid produce identical token sequences, regardless of differences in sampling rate or point density that would yield distinct sequences under other representations. This improves generalization across diverse input conditions.

\paragraph{Compression.} While the base vocabulary is small, raw ScribeTokens sequences can be long due to the pixel-level granularity of the decomposition. We apply BPE to compress sequences, with merges restricted to direction tokens only---pen-state tokens \texttt{[UP]} and \texttt{[DOWN]} are never merged---ensuring that stroke boundaries remain explicit. Since any merged token can always be decomposed back into its constituent base direction tokens, the representation remains OOV-free by construction.

\paragraph{Detokenization.} For generation tasks, the predicted token sequence must be detokenized back to ink coordinates. Starting from an origin, each direction token specifies a unit step, and pen-state tokens delimit stroke boundaries. The recovered integer coordinates are scaled by $\delta$ to return to the original coordinate space, and Savitzky--Golay smoothing~\citep{savitzky1964smoothing} is applied to mitigate staircase artifacts from the grid discretization (Appendix~\ref{app:quantization}).

\subsection{Task Formulation}

We formulate all tasks under a unified prompt-completion framework. Given a prompt sequence~$\mathbf{x}$ and a completion sequence~$\mathbf{y} = (y_1, \ldots, y_T)$, a causal model is trained to maximize the conditional log-likelihood:
\begin{equation*}
  \log p(\mathbf{y} \mid \mathbf{x}) = \sum_{t=1}^{T} \log p(y_t \mid \mathbf{x},\, y_{<t}).
\end{equation*}
At inference, $\mathbf{y}$ is decoded autoregressively from a beginning-of-sequence state until an end-of-sequence condition is reached.

Let $\mathbf{s}$ denote the sequence encoding of a digital ink~$\mathcal{I}$ under the chosen representation, and let $\mathbf{c}$ denote its text transcript. The three tasks considered in this work---next-token prediction (NTP), handwritten text recognition (HTR), and handwritten text generation (HTG)---are special cases of this framework, differing only in what constitutes $\mathbf{x}$ and $\mathbf{y}$:

\begin{itemize}
  \item NTP: $\mathbf{x} = \varnothing$, $\mathbf{y} = \mathbf{s}$ --- unconditional ink generation as a self-supervised pretraining objective.
  \item HTR: $\mathbf{x} = \mathbf{s}$, $\mathbf{y} = \mathbf{c}$ --- the model reads ink and produces a text transcription~$\mathbf{c}$.
  \item HTG: $\mathbf{x} = \mathbf{c}$, $\mathbf{y} = \mathbf{s}$ --- the model generates ink conditioned on a text prompt.
\end{itemize}

\paragraph{Training loss.} The choice of loss depends on the completion modality. When $\mathbf{y}$ consists of discrete tokens---text for HTR, or a token representation of ink for NTP and HTG---we use standard cross-entropy loss. When $\mathbf{y}$ uses a vector representation of ink, coordinates are modeled via a mixture density network and pen states via a categorical distribution, with the total loss being the sum of the negative log-likelihood for coordinates and cross-entropy for pen states~\citep{graves2013generating}.
\section{Experiments}
\label{sec:experiments}

\subsection{Datasets}
\paragraph{IAM.} The IAM On-Line Handwriting Database (IAM-OnDB)~\citep{liwicki2005iam} is a widely adopted benchmark for digital ink analysis, containing labeled text lines from 221 writers. We use the standard writer-disjoint split, which---after removing corrupt samples---yields 5{,}042 train, 2{,}264 validation, and 3{,}541 test text lines. Its samples are full text lines, so sequences are long, and the dataset is relatively small---making IAM a challenging setting for evaluating long-range modeling and data efficiency.

\paragraph{DeepWriting.} The DeepWriting dataset~\citep{aksan2018deepwriting} is primarily derived from IAM-OnDB by filtering low-quality writers and segmenting text lines into individual words. We use only the IAM-derived portion, excluding supplementary collections that differ substantially in ink scale. As no standardized split exists, we use a random 80/10/10 train/validation/test partition without writer-disjoint constraints, yielding 36{,}912, 4{,}614, and 4{,}614 samples respectively. Results on DeepWriting therefore reflect interpolation within observed writing styles rather than generalization to unseen writers. Compared to IAM, DeepWriting offers shorter sequences and more training data, providing a complementary regime for evaluation.

\subsection{Tokenization Analysis}
We train BPE tokenizers for AbsTokens, RelTokens, TextTokens, and ScribeTokens on the IAM training set across different quantization parameters $\delta$. We use IAM rather than DeepWriting because its line-level samples contain long-range pen movements---such as spaces between words---that are absent from word-level data, yielding tokenizers that generalize better to diverse digital ink. Compression ratios and OOV rates are measured on the IAM validation set; some AbsTokens and RelTokens configurations are absent where the base vocabulary already exceeds the budget, leaving no capacity for BPE merges.

\begin{figure}[t]
  \centering
  \includegraphics[width=\textwidth]{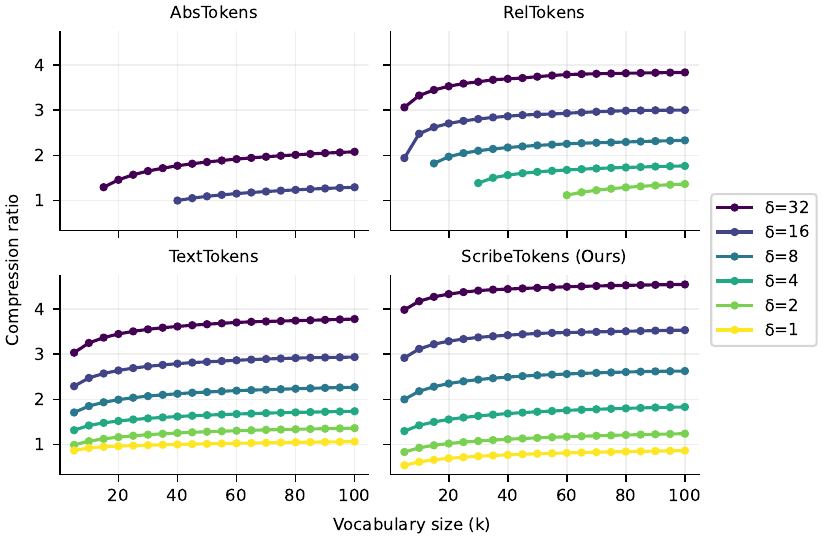}
  \caption{Average compression ratios ($\uparrow$) of BPE-based digital ink representations on the IAM validation set, across target vocabulary sizes and quantization parameters~$\delta$. ScribeTokens consistently achieves the highest compression across nearly all settings.}
  \label{fig:compression}
\end{figure}

\paragraph{Compression.} Figure~\ref{fig:compression} shows average compression ratios across target vocabulary sizes and quantization parameters. ScribeTokens consistently achieves the highest compression across all settings, except at very fine quantization ($\delta \in \{1, 2\}$).

\begin{figure}[t]
  \centering
  \includegraphics[width=\textwidth]{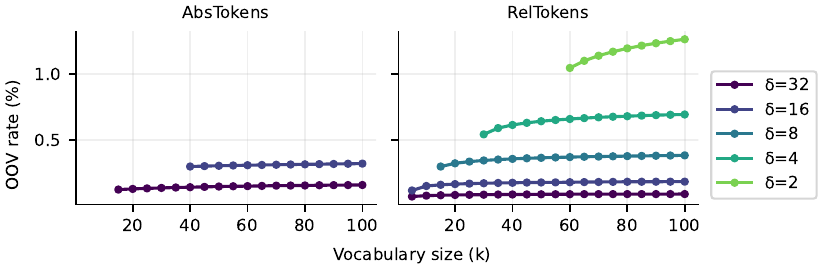}
  \caption{Average out-of-vocabulary (OOV) rates ($\downarrow$) of BPE-based digital ink representations on the IAM validation set, across target vocabulary sizes and quantization parameters~$\delta$. ScribeTokens and TextTokens are OOV-free by construction, while AbsTokens and RelTokens exhibit non-zero OOV rates as their coordinate-based vocabularies inevitably encounter unseen values at test time.}
  \label{fig:oov}
\end{figure}

\paragraph{OOV.} Figure~\ref{fig:oov} shows average OOV rates across vocabulary sizes and quantization parameters. ScribeTokens and TextTokens are OOV-free by construction. AbsTokens exhibits OOV rates of 0.15--0.3\%, while RelTokens ranges from approximately 0.1\% up to over 1.2\%. These rates are non-trivial given that typical IAM ink samples contain on the order of 1{,}000 points.

\paragraph{Quantization artifacts.} Beyond compression and OOV behavior, $\delta$ also governs the fidelity of the reconstructed ink. To evaluate this qualitatively, we quantize IAM validation samples at different values of $\delta$ and apply Savitzky--Golay post-processing (Appendix~\ref{app:quantization}). Figure~\ref{fig:discretization} shows an example: at $\delta \leq 8$, post-processed inks are visually indistinguishable from the original; artifacts become noticeable around $\delta = 16$ and progressively degrade, with ink becoming unrecognizable at $\delta \geq 128$.

\subsection{Downstream Setup}
\label{sec:setup}

Based on considerations of compression ratio, OOV rate, and reconstruction quality, we fix $\delta = 8$ and a target vocabulary size of 32{,}000 for all downstream experiments. Four representations are compared: Point-5, RelTokens, TextTokens, and ScribeTokens. Point-3 is excluded as it cannot signal end of sequence, and AbsTokens because its base vocabulary already exceeds the vocabulary budget, leaving no capacity for BPE merges. The same model architecture and training is used for all representations and tasks, ensuring that performance differences are driven by the choice of representation.

\paragraph{Architecture and training.} Our model is a 12-layer decoder-only Transformer~\citep{vaswani2017attention} following the LLaMA design~\citep{touvron2023llama}, with approximately 34M parameters for token-based models. Point-5, which replaces the large embedding table with lightweight linear projections, has approximately 21M parameters; further details are given in Appendix~\ref{app:point5}. Models are trained with AdamW~\citep{loshchilov2019decoupled} for up to 200 epochs with early stopping, using stochastic geometric augmentations on ink inputs. Full architecture and training details are provided in Appendix~\ref{app:architecture}.

\subsection{Handwritten Text Recognition}
\label{sec:htr}

The task is to predict the text transcript from an input ink sequence. Models denoted +PT are first pretrained with next-ink-token prediction on the training set, then fine-tuned on the same training set for recognition. Table~\ref{tab:htr} reports character error rate (CER) and exact-match accuracy under autoregressive decoding (temperature 0).

\begin{table}[t]
  \caption{Handwritten text recognition on IAM and DeepWriting, measured by CER ($\downarrow$) and accuracy ($\uparrow$) under autoregressive decoding (temperature 0). +PT denotes initialization from next-ink-token prediction pretraining prior to task fine-tuning. Deltas show the effect of pretraining: {\scriptsize\textcolor{teal}{teal}} indicates improvement, {\scriptsize\textcolor{red}{red}} degradation. Best results are \textbf{bolded}.}
  \label{tab:htr}
  \begin{center}
    \begin{tabular}{ll
    S[table-format=2.2,
    table-space-text-post={\,{\scriptsize\textcolor{red}{(+00.00)}}}]
    S[table-format=2.2,
    table-space-text-post={\,{\scriptsize\textcolor{red}{(+00.00)}}}]
    S[table-format=2.2,
    table-space-text-post={\,{\scriptsize\textcolor{red}{(+00.00)}}}]
    S[table-format=2.2,
  table-space-text-post={\,{\scriptsize\textcolor{red}{(+00.00)}}}]}
  \toprule
  &  & \multicolumn{2}{c}{IAM} & \multicolumn{2}{c}{DeepWriting} \\
  \cmidrule(lr){3-4} \cmidrule(lr){5-6}
  Method &  & {CER (\%) $\downarrow$} & {Acc (\%) $\uparrow$} & {CER
  (\%) $\downarrow$} & {Acc (\%) $\uparrow$} \\
  \midrule
  Point-5 &  & 9.43 & 31.38 & 10.94 & 81.97 \\
  & +PT & 13.63\,{\scriptsize\textcolor{red}{(+4.19)}} &
  18.92\,{\scriptsize\textcolor{red}{(-12.45)}} &
  10.25\,{\scriptsize\textcolor{teal}{(-0.69)}} &
  83.16\,{\scriptsize\textcolor{teal}{(+1.19)}} \\
  \addlinespace
  RelTokens &  & 12.69 & 25.11 & 11.22 & 81.60 \\
  & +PT & 9.16\,{\scriptsize\textcolor{teal}{(-3.53)}} &
  30.75\,{\scriptsize\textcolor{teal}{(+5.65)}} &
  10.49\,{\scriptsize\textcolor{teal}{(-0.73)}} &
  82.73\,{\scriptsize\textcolor{teal}{(+1.13)}} \\
  \addlinespace
  TextTokens &  & 82.00 & 0.00 & 11.65 & 81.17 \\
  & +PT & 9.54\,{\scriptsize\textcolor{teal}{(-72.46)}} &
  29.57\,{\scriptsize\textcolor{teal}{(+29.57)}} &
  10.07\,{\scriptsize\textcolor{teal}{(-1.58)}} &
  83.33\,{\scriptsize\textcolor{teal}{(+2.17)}} \\
  \midrule
  ScribeTokens (Ours) &  & 13.15 & 22.79 & 10.75 & 82.29 \\
  & +PT & \bfseries 8.27
  \mdseries\,{\scriptsize\textcolor{teal}{(-4.87)}} & \bfseries 32.93
  \mdseries\,{\scriptsize\textcolor{teal}{(+10.14)}} & \bfseries 9.83
  \mdseries\,{\scriptsize\textcolor{teal}{(-0.92)}} & \bfseries 83.40
  \mdseries\,{\scriptsize\textcolor{teal}{(+1.11)}} \\
  \bottomrule
\end{tabular}

  \end{center}
\end{table}

\paragraph{No pretraining.} ScribeTokens leads on DeepWriting (10.75\% CER), the only token-based model to outperform Point-5 (10.94\% CER). On IAM, Point-5 leads (9.43\% CER), outperforming all token representations. TextTokens fails entirely on IAM (82.00\% CER, 0\% accuracy). ScribeTokens and RelTokens models exhibit double descent during training, but TextTokens never escapes the ascent (Appendix~\ref{app:double}); attention analysis suggests it collapses into a language model (Appendix~\ref{app:attention}).

\paragraph{With pretraining.} ScribeTokens + PT achieves the best results on both datasets (8.27\% CER on IAM; 9.83\% CER on DeepWriting). Pretraining consistently improves all token-based models, with especially large gains on IAM where data is scarce---TextTokens recovers from complete failure (82.00\% $\to$ 9.54\% CER). Point-5 is the exception---pretraining \emph{degrades} its IAM performance (9.43\% $\to$ 13.63\% CER); we hypothesize why in Appendix~\ref{app:pretraining}. For reference, \citet{graves2009novel} report 11.5\% CER on IAM with a bidirectional LSTM, CTC, and no language model; state-of-the-art systems reach below 5\% CER with language models---a choice orthogonal to ink representation.

\subsection{Handwritten Text Generation}
\label{sec:htg}

The task is to produce an ink sequence conditioned on a text prompt. Table~\ref{tab:htg} reports CER and exact-match accuracy under autoregressive decoding (temperature 1), where generated inks are scored against the text prompt by our best-performing recognizer---ScribeTokens + PT---trained exclusively on real handwriting. All outputs are first detokenized to raw ink coordinates and post-processed (Appendix~\ref{app:quantization}). Sampling invariance (Section~\ref{sec:scribetokens}) then ensures recognition depends only on the rasterized path. Using a ScribeTokens-trained recognizer therefore does not bias evaluation toward ScribeTokens-generated ink: indeed, TextTokens matches ScribeTokens' CER on IAM with pretraining (Table~\ref{tab:htg}). Qualitative examples of generated inks are provided in Appendix~\ref{app:htg_samples}.

\begin{table}[t]
  \caption{Handwritten text generation on IAM and DeepWriting, measured by CER ($\downarrow$) and accuracy ($\uparrow$) under autoregressive decoding (temperature 1). +PT denotes initialization from next-ink-token prediction pretraining prior to task fine-tuning. Generated inks are post-processed (Appendix~\ref{app:quantization}) and evaluated by the best HTR model (ScribeTokens + PT; Table~\ref{tab:htr}). Deltas show the effect of pretraining: {\scriptsize\textcolor{teal}{teal}} indicates improvement, {\scriptsize\textcolor{red}{red}} degradation. Best results are \textbf{bolded}.}
  \label{tab:htg}
  \begin{center}
    \begin{tabular}{ll
    S[table-format=2.2,
    table-space-text-post={\,{\scriptsize\textcolor{red}{(+00.00)}}}]
    S[table-format=2.2,
    table-space-text-post={\,{\scriptsize\textcolor{red}{(+00.00)}}}]
    S[table-format=2.2,
    table-space-text-post={\,{\scriptsize\textcolor{red}{(+00.00)}}}]
    S[table-format=2.2,
  table-space-text-post={\,{\scriptsize\textcolor{red}{(+00.00)}}}]}
  \toprule
  &  & \multicolumn{2}{c}{IAM} & \multicolumn{2}{c}{DeepWriting} \\
  \cmidrule(lr){3-4} \cmidrule(lr){5-6}
  Method &  & {CER (\%) $\downarrow$} & {Acc (\%) $\uparrow$} & {CER
  (\%) $\downarrow$} & {Acc (\%) $\uparrow$} \\
  \midrule
  Point-5 &  & 70.29 & 0.03 & 14.36 & 71.09 \\
  & +PT & 16.83\,{\scriptsize\textcolor{teal}{(-53.46)}} &
  15.96\,{\scriptsize\textcolor{teal}{(+15.93)}} &
  26.84\,{\scriptsize\textcolor{red}{(+12.48)}} &
  51.24\,{\scriptsize\textcolor{red}{(-19.85)}} \\
  \addlinespace
  RelTokens &  & 25.89 & 4.43 & \bfseries 12.75 \mdseries & \bfseries
  73.04 \mdseries \\
  & +PT & 11.93\,{\scriptsize\textcolor{teal}{(-13.96)}} &
  20.33\,{\scriptsize\textcolor{teal}{(+15.90)}} &
  12.98\,{\scriptsize\textcolor{red}{(+0.23)}} &
  72.48\,{\scriptsize\textcolor{red}{(-0.56)}} \\
  \addlinespace
  TextTokens &  & 13.65 & 16.58 & 15.32 & 70.85 \\
  & +PT & 10.45\,{\scriptsize\textcolor{teal}{(-3.20)}} &
  22.71\,{\scriptsize\textcolor{teal}{(+6.13)}} &
  14.12\,{\scriptsize\textcolor{teal}{(-1.20)}} &
  72.61\,{\scriptsize\textcolor{teal}{(+1.76)}} \\
  \midrule
  ScribeTokens (Ours) &  & 17.33 & 12.93 & 15.47 & 72.78 \\
  & +PT & \bfseries 10.45
  \mdseries\,{\scriptsize\textcolor{teal}{(-6.88)}} & \bfseries 23.84
  \mdseries\,{\scriptsize\textcolor{teal}{(+10.90)}} &
  16.02\,{\scriptsize\textcolor{red}{(+0.55)}} &
  68.23\,{\scriptsize\textcolor{red}{(-4.55)}} \\
  \bottomrule
\end{tabular}

  \end{center}
\end{table}

\paragraph{No pretraining.} On DeepWriting, all methods perform reasonably, with RelTokens leading (12.75\% CER). On IAM, where sequences are long, all methods struggle---TextTokens fares best (13.65\% CER) but Point-5 nearly fails entirely (70.29\% CER), likely because the lack of compression yields excessively long sequences. RelTokens' contrasting performance across datasets---strong on word-level DeepWriting but weak on sentence-level IAM (25.89\% CER)---is consistent with its OOV susceptibility: sentence-level samples contain large inter-word jumps that produce rare displacements unlikely to appear in the vocabulary.

\paragraph{With pretraining.} ScribeTokens + PT achieves the best IAM results (10.45\% CER, 23.84\% accuracy), with TextTokens + PT closely following (10.45\% CER, 22.71\% accuracy). Pretraining dramatically improves all methods on IAM, including ScribeTokens (17.33\% $\to$ 10.45\% CER) and Point-5 (70.29\% $\to$ 16.83\% CER). On DeepWriting, pretraining effects are mixed, and no pretrained model surpasses RelTokens trained from scratch (12.75\% CER). Pretraining benefits are most pronounced on IAM, where data is scarce and sequences are long; on the larger, word-level DeepWriting set, task supervision alone may suffice to learn good representations.

\subsection{Next-Ink-Token Prediction}
\label{sec:ntp}

A single pretrained checkpoint initializes both recognition and generation. Having shown that this task-agnostic pretraining improves final performance in both settings, we briefly examine its practical and mechanistic effects.

\paragraph{Speedup.} Beyond final performance, pretraining accelerates convergence by up to \SI{21}{\x} on HTR and \SI{83}{\x} on HTG, measured by the number of fine-tuning epochs needed to reach the baseline converged loss (Appendix~\ref{app:speedup}). These gains are most pronounced in the low-data regime (IAM) and more modest on the larger DeepWriting set.

\paragraph{Intuition.} Token embeddings are initialized randomly and carry no spatial information---a cold-start problem absent from continuous vectors. Next-ink-token prediction forces the model to resolve this (Appendix~\ref{app:pretraining}). Attention analysis corroborates the effect---pretraining shifts the fraction of attention allocated to ink from as low as 50.7\% to as high as 91.2\%, indicating that pretrained models decode primarily from the ink signal rather than memorized text patterns (Appendix~\ref{app:attention}).

\section{Conclusion}
\label{sec:conclusion}

We introduced ScribeTokens, a canonical tokenization of digital ink that decomposes pen strokes into unit directional steps via Bresenham's line algorithm. A fixed base vocabulary of just 10 tokens suffices to encode any ink, eliminating the out-of-vocabulary issues, large vocabulary sizes, and fragile syntax of prior tokenizations while remaining invariant to sampling rate. BPE over this base vocabulary yields aggressive compression without sacrificing the OOV-free guarantee.

Our experiments revealed that token representations are far more effective than vectors for generation, and that ScribeTokens is the only token representation to outperform vectors on recognition without pretraining---resolving a key limitation of prior tokenizations. We further showed that next-ink-token prediction is an effective self-supervised pretraining strategy for ink tokens, accelerating convergence by up to \SI{83}{\x} and yielding consistent gains in data-limited settings. With pretraining, ScribeTokens achieved the best recognition results on both datasets and the best generation results on IAM.

\paragraph{Limitations and future work.} We evaluated only on English handwriting recognition and generation with a single 34M-parameter decoder-only Transformer; generalization to other scripts, tasks, and model families remains to be verified. Results are from single runs. Additionally, next-ink-token prediction pretraining was performed on the same training sets rather than large unlabeled corpora; we expect the mixed generation results on DeepWriting to diminish when pretraining and fine-tuning use separate datasets. Scaling pretraining data and efficient long-sequence attention are directly applicable. ScribeTokens provides the vocabulary; the language modeling playbook provides the roadmap.

\begin{ack}
  The author thanks Fr\'{e}d\'{e}ric Fillieux, Huidong Liang, and Zeli Wang for feedback on earlier drafts. The author also thanks his family for their support. This work was not supported by any external grants or organizations.
\end{ack}

{

}

\newpage
\appendix

\section{Quantization Artifacts}
\label{app:quantization}

\begin{figure}[t]
  \centering
  \includegraphics[width=\textwidth]{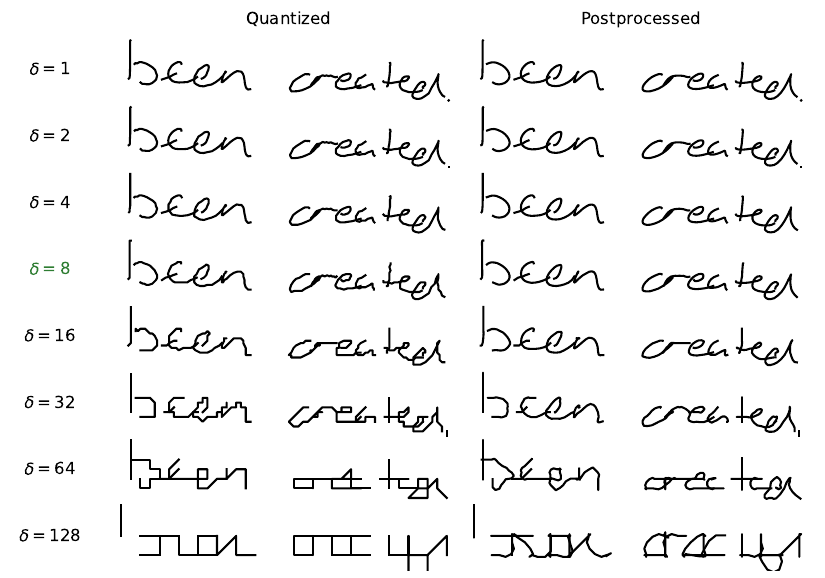}
  \caption{Effect of quantization parameter~$\delta$ on reconstruction quality. Each row shows the same IAM sample quantized at a different $\delta$, displayed both as raw quantized ink (left) and after Savitzky--Golay post-processing (right). Post-processed inks are visually indistinguishable from the original for $\delta \leq 8$; the row at $\delta = 8$ (highlighted in green) maximizes compression without sacrificing fidelity and is used in all downstream experiments.}
  \label{fig:discretization}
\end{figure}

Quantizing ink coordinates to a discrete grid introduces staircase artifacts in reconstructed strokes. Figure~\ref{fig:discretization} visualizes these artifacts across eight values of~$\delta$, both before and after the post-processing described below.

\paragraph{Smoothing.} To recover smooth trajectories, we apply Savitzky--Golay filtering~\citep{savitzky1964smoothing} independently to the $x$- and $y$-coordinate sequences of each stroke. We prefer Savitzky--Golay over Gaussian smoothing because its local polynomial fitting better preserves peaks and sharp transitions in the stroke geometry. We use a window size of $w=7$ and polynomial order $k=3$ for all experiments.

\paragraph{Downsampling.} For ScribeTokens, Bresenham's line algorithm introduces many intermediate pixel-level points, producing denser strokes than the original ink. To counteract this, we downsample each generated stroke (retaining every $d$-th point) before applying the filter. The optimal downsampling rate $d$ depends on the choice of $\delta$. Since generated coordinates are rescaled by $\delta$, larger values of $\delta$ produce strokes with fewer points and require smaller $d$. We use $d=2$ for all experiments ($\delta=8$).

\section{Architecture and Training}
\label{app:architecture}

Table~\ref{tab:hyperparameters} lists all architecture and training hyperparameters used across experiments. Since RelTokens models are susceptible to out-of-vocabulary tokens at inference, we inject \texttt{[UNKNOWN]} tokens into its training data with probability 0.4\%, matched to the empirical OOV rate on the validation set, so the model learns to handle unseen tokens gracefully. All models are implemented in PyTorch. All experiments are run on a single NVIDIA GH200 GPU; total compute across all runs is approximately 85 GPU-hours.

\begin{table}[t]
  \centering
  \caption{Architecture and training hyperparameters shared across all representations and tasks.}
  \label{tab:hyperparameters}
  \begin{tabular}{ll}
  \toprule
  \multicolumn{2}{l}{\textbf{Architecture}} \\
  \midrule
  Model type & Decoder-only Transformer (LLaMA-style) \\
  Parameters & ${\sim}\num{34}$M (token-based) / ${\sim}\num{21}$M (Point-5) \\
  Layers & \num{12} \\
  Attention heads & \num{6} \\
  Hidden dimension & \num{384} \\
  Positional encoding & RoPE~\citep{su2024roformer} \\
  Activation & SwiGLU~\citep{shazeer2020glu} (expansion factor $\num{8}/\num{3}$) \\
  Normalization & Pre-norm RMSNorm~\citep{zhang2019root} \\
  Embedding tying & Yes (token-based models) \\
  Dropout & \num{0.2} \\
  \midrule
  \multicolumn{2}{l}{\textbf{Optimization}} \\
  \midrule
  Optimizer & AdamW \\
  Learning rate & \num{3e-4} (constant) \\
  Weight decay & \num{0.1} \\
  Max gradient norm & \num{1.0} \\
  Batch size & \num{64} (DeepWriting) / \num{32} (IAM) \\
  Mixed precision & bfloat16 \\
  Early stopping & \num{200} epochs, patience \num{50} \\
  \midrule
  \multicolumn{2}{l}{\textbf{Data augmentation} (each applied independently with $p=\num{0.5}$)} \\
  \midrule
  Random scaling & $\pm\SI{30}{\percent}$ \\
  Shearing & $\pm\num{0.5}$ \\
  Rotation & $\pm\ang{5}$ \\
  Gaussian jitter & $\sigma = \num{5}$ \\
  \bottomrule
\end{tabular}
\end{table}

\section{Point-5 Model}
\label{app:point5}

The Point-5 representation encodes each time step as a 5-dimensional vector $(\Delta x, \Delta y, p_1, p_2, p_3)$, which is projected to the model's hidden dimension via a learned linear layer. All ink coordinates are scaled down by a factor of 10 to keep offset magnitudes in a stable range. A fixed beginning-of-sequence state $(0, 0, 0, 1, 0)$, corresponding to zero offset in the pen-up state, is prepended to the sequence to initiate generation. For generation, the model must produce both continuous coordinates and discrete pen states, requiring a specialized generation head and several numerical safeguards.

\paragraph{Generation head.} The generation head models coordinates and pen states separately. For coordinates, we use a mixture density network (MDN) with $K=20$ mixture components. The hidden state is mapped to mixture weights $\pi_k \in (0,1)$ (linear + softmax), means $\boldsymbol{\mu}_k \in \mathbb{R}^2$ (linear), standard deviations $\boldsymbol{\sigma}_k \in \mathbb{R}^2$ (linear + softplus), and correlation coefficients $\rho_k \in (-1,1)$ (linear + tanh) for each component. Separately, the pen state $(p_1, p_2, p_3)$ is mapped from the hidden state to a categorical distribution (linear + softmax).

\paragraph{Numerical safeguards.} The MDN head requires several safeguards to train reliably. Standard deviations are clamped to a minimum of $\sigma_{\min} = 0.1$ to prevent the mixture components from collapsing to near-zero variance, and correlation coefficients are bounded to $|\rho_k| \leq 0.99$ to avoid degenerate covariance matrices. Without these measures, training diverges early. These constraints are unnecessary for token-based models, which use standard cross-entropy loss.

\section{Pretraining Speedup}
\label{app:speedup}

Beyond final task performance, next-ink-token prediction pretraining also accelerates downstream convergence. Tables~\ref{tab:convergence-htr} and~\ref{tab:convergence-htg} report, for each representation, the number of pretraining epochs run, the number of epochs to converge without pretraining, and how many fine-tuning epochs the pretrained model needs to reach the same converged validation loss.

\paragraph{HTR.} On IAM, ScribeTokens achieves the largest speedup at \SI{21.4}{\x}: the baseline converges in 193 epochs, while the pretrained model reaches the same loss in just 9 fine-tuning epochs. Notably, the total number of epochs for ScribeTokens with pretraining (90 pretraining + 9 fine-tuning = 99 epochs) is fewer than training without pretraining (193 epochs), while also achieving substantially better final performance (8.27\% vs.\ 13.15\% CER).

\begin{table}[t]
  \centering
  \caption{Convergence speedup from pretraining for HTR. \textit{PT Ep.}: pretraining epochs run. \textit{No PT}: epochs to converge without pretraining. \textit{+PT}: fine-tuning epochs to reach the same converged loss. \textit{Spd.}: speedup ratio (No PT\,/\,+PT). ``--'' indicates the pretrained model never reached the baseline loss. Best speedups are \textbf{bolded}.}
  \label{tab:convergence-htr}
  \begin{tabular}{l
    S[table-format=2] S[table-format=3] S[table-format=2]
    S[table-format=2.1, table-space-text-post=$\times$]
    S[table-format=3] S[table-format=2] S[table-format=1]
  S[table-format=1.1, table-space-text-post=$\times$]}
  \toprule
  & \multicolumn{4}{c}{IAM} & \multicolumn{4}{c}{DeepWriting} \\
  \cmidrule(lr){2-5} \cmidrule(lr){6-9}
  Method & {PT Ep.} & {No PT} & {+PT} & {Spd.} & {PT Ep.} & {No PT} &
  {+PT} & {Spd.} \\
  \midrule
  Point-5 & 36 & 171 & {--} & {--} & 55 & 13 & {--} & {--} \\
  RelTokens & 72 & 193 & 13 & 14.8 $\times$ & 99 & 23 & 9 & 2.6 $\times$ \\
  TextTokens & 99 & 18 & 6 & 3.0 $\times$ & 100 & 30 & 6 & \bfseries
  5.0 $\times$ \\
  \midrule
  ScribeTokens (Ours) & 90 & 193 & 9 & \bfseries 21.4 $\times$ & 99 &
  26 & 8 & 3.2 $\times$ \\
  \bottomrule
\end{tabular}

\end{table}

\paragraph{HTG.} The speedups for HTG are even more pronounced. On IAM, ScribeTokens reaches the baseline loss in a single fine-tuning epoch, yielding an \SI{83.0}{\x} speedup. Interestingly, although pretraining worsens HTG task metrics for some models on DeepWriting (Table~\ref{tab:htg}), every pretrained token-based model still achieves lower cross-entropy loss than its non-pretrained counterpart. This demonstrates a disconnect between cross-entropy loss and generation quality.

\begin{table}[t]
  \centering
  \caption{Convergence speedup from pretraining for HTG. \textit{PT Ep.}: pretraining epochs run. \textit{No PT}: epochs to converge without pretraining. \textit{+PT}: fine-tuning epochs to reach the same converged loss. \textit{Spd.}: speedup ratio (No PT\,/\,+PT). ``--'' indicates the pretrained model never reached the baseline loss. Best speedups are \textbf{bolded}.}
  \label{tab:convergence-htg}
  \begin{tabular}{l
    S[table-format=2] S[table-format=2] S[table-format=2]
    S[table-format=2.1, table-space-text-post=$\times$]
    S[table-format=3] S[table-format=2] S[table-format=2]
  S[table-format=1.1, table-space-text-post=$\times$]}
  \toprule
  & \multicolumn{4}{c}{IAM} & \multicolumn{4}{c}{DeepWriting} \\
  \cmidrule(lr){2-5} \cmidrule(lr){6-9}
  Method & {PT Ep.} & {No PT} & {+PT} & {Spd.} & {PT Ep.} & {No PT} &
  {+PT} & {Spd.} \\
  \midrule
  Point-5 & 36 & 41 & 20 & 2.0 $\times$ & 55 & 61 & {--} & {--} \\
  RelTokens & 72 & 80 & 1 & 80.0 $\times$ & 99 & 98 & 18 & \bfseries
  5.4 $\times$ \\
  TextTokens & 99 & 87 & 4 & 21.8 $\times$ & 100 & 99 & 19 & 5.2 $\times$ \\
  \midrule
  ScribeTokens (Ours) & 90 & 83 & 1 & \bfseries 83.0 $\times$ & 99 &
  96 & 27 & 3.6 $\times$ \\
  \bottomrule
\end{tabular}

\end{table}

\section{Double Descent}
\label{app:double}

\begin{figure}[t]
  \centering
  \includegraphics[width=0.618\textwidth]{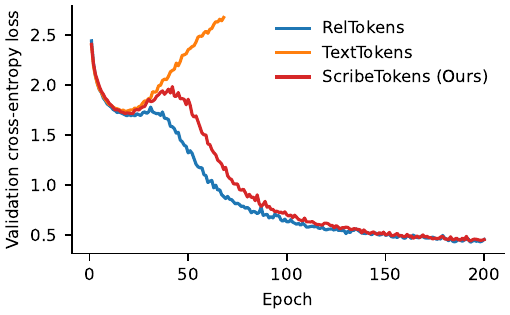}
  \caption{Validation cross-entropy loss during HTR training on IAM without pretraining. RelTokens and ScribeTokens exhibit a double descent pattern, recovering after an initial loss increase, while TextTokens diverges and fails to learn.}
  \label{fig:double_descent}
\end{figure}

For HTR on IAM, TextTokens without pretraining achieves 82.00\% CER and 0\% accuracy (Table~\ref{tab:htr}), effectively failing to learn the task. Figure~\ref{fig:double_descent} plots the validation cross-entropy loss over training for the three token-based models. All three initially decrease before rising around epoch 20--40, but only RelTokens and ScribeTokens recover and descend to convergence---exhibiting a double descent pattern. TextTokens continues to diverge, never escaping the ascent.

\section{Attention Analysis}
\label{app:attention}

We hypothesize that the double descent in Appendix~\ref{app:double} reflects a collapse of the recognition model into a language model: when the ink representation is difficult to leverage, the model overfits to text transcripts rather than learning to read the ink. To test this, we visualize how the most recently decoded character token attends to the ink input using attention rollout~\citep{abnar2020quantifying}, which collapses attention weights from all layers into a single heatmap. We also report the \emph{attention split}: the fraction of total rolled-out attention mass allocated to ink tokens versus the previously decoded text prefix. Across all three models examined below, we find that recognition performance correlates strongly with the fraction of attention allocated to ink.

\begin{figure}[t]
  \centering
  \includegraphics[width=0.618\textwidth]{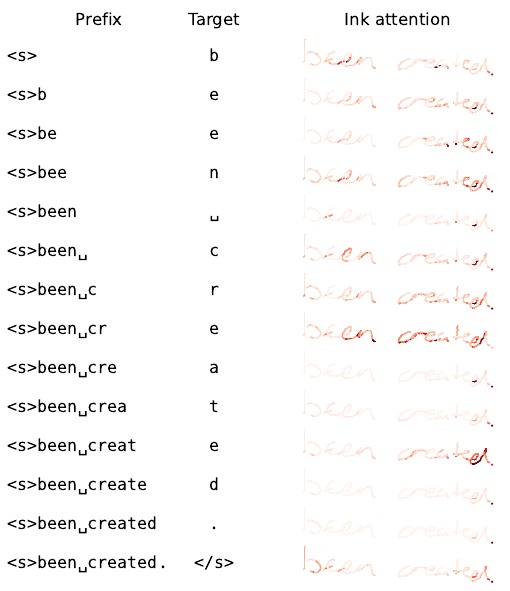}
  \caption{Attention rollout for the TextTokens HTR model (82.00\% CER, 0\% accuracy on IAM). Each subplot shows the rolled-out attention from the most recently decoded character token onto the ink input at a different decoding step. The attention is diffuse with no discernable correspondence to the target character. Attention split: 50.7\% ink, 49.3\% text.}
  \label{fig:attn_htr_text}
\end{figure}

\paragraph{TextTokens.} Figure~\ref{fig:attn_htr_text} shows the attention rollout for the TextTokens HTR model, which failed to learn the task (82.00\% CER, 0\% accuracy on IAM). The attention is diffuse with no discernable correspondence to the target character. The attention split is 50.7\% ink and 49.3\% text. Unlike ScribeTokens and RelTokens, whose tokens directly encode pen movements, TextTokens' lexical vocabulary (digits, minus signs, spaces) carries no geometric meaning, requiring the model to parse numeral sequences before it can recover spatial structure. This makes useful information harder to extract from the ink signal.

\begin{figure}[t]
  \centering
  \includegraphics[width=0.618\textwidth]{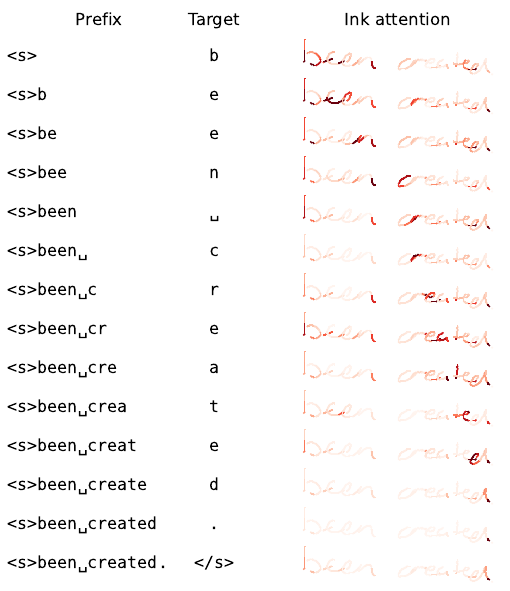}
  \caption{Attention rollout for the ScribeTokens HTR model (13.15\% CER, 22.79\% accuracy on IAM). Compared to TextTokens (Figure~\ref{fig:attn_htr_text}), the attention exhibits a clear left-to-right pattern, concentrating slightly ahead of the character being recognized, consistent with causal lookahead. Attention split: 66.9\% ink, 33.1\% text.}
  \label{fig:attn_htr_scribe}
\end{figure}

\paragraph{ScribeTokens.} By contrast, Figure~\ref{fig:attn_htr_scribe} shows the same visualization for the ScribeTokens HTR model, which converged without pretraining (13.15\% CER, 22.79\% accuracy on IAM). Although some spurious attention remains, there is a clear left-to-right pattern: attention concentrates slightly ahead of the character being recognized, consistent with the lookahead expected of a causal decoder. The attention split shifts to 66.9\% ink and 33.1\% text. Unlike TextTokens, both ScribeTokens and RelTokens use spatially grounded vocabularies whose tokens directly encode pen movements, making the ink signal easier to leverage.

\paragraph{ScribeTokens + PT.} The best-performing HTR model, ScribeTokens + PT (8.27\% CER, 32.93\% accuracy), pushes the attention split further to 91.2\% ink and 8.8\% text. Interestingly, although the model attends overwhelmingly to ink, the rolled-out attention loses all spatial localization, making the heatmaps uninterpretable. The progression across all three models is nonetheless consistent: as the fraction of attention allocated to ink increases (50.7\% $\to$ 66.9\% $\to$ 91.2\%), recognition performance improves correspondingly. While this analysis is correlational, it supports the hypothesis that TextTokens' failure stems from an inability to effectively leverage the ink signal.

\begin{figure}[t]
  \centering
  \includegraphics[width=0.618\textwidth]{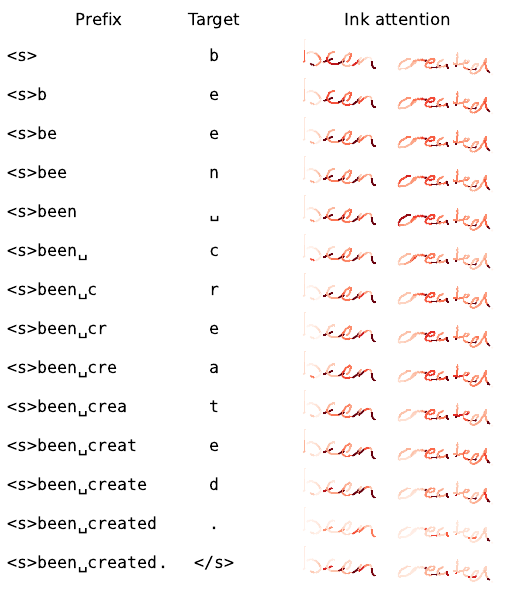}
  \caption{Attention rollout for the ScribeTokens HTR model with pretraining (8.27\% CER, 32.93\% accuracy on IAM). The attention loses all spatial localization, making the heatmaps uninterpretable. Compared to the non-pretrained model (Figure~\ref{fig:attn_htr_scribe}; 66.9\% ink, 33.1\% text), pretraining shifts attention almost entirely to ink. Attention split: 91.2\% ink, 8.8\% text.}
  \label{fig:attn_htr_sft_scribe}
\end{figure}

\section{Understanding Pretraining}
\label{app:pretraining}

Token-based models require learned embeddings that map discrete tokens to continuous vectors. At initialization, these embeddings are random and carry no spatial meaning, so the model must first discover which tokens correspond to similar pen movements before it can reason about stroke geometry. By contrast, Point-5's continuous coordinates inherently encode spatial proximity, giving the model a useful inductive bias. Next-ink-token prediction pretraining addresses this cold-start problem: to predict the next stroke segment, the model must learn embeddings that capture the spatial relationships between tokens.

\paragraph{HTR.} Success on next-ink-token prediction requires an implicit ability to recognize what has already been written: the model must track context to decide what comes next. For example, completing the word ``the'' after writing ``th'' requires recognizing the previously generated characters. Pretraining thus builds internal representations that capture character identity from ink patterns---precisely the capability HTR requires. Pretraining also bootstraps the model into relying on the ink signal rather than the text prefix. As shown in Appendix~\ref{app:double}, the non-pretrained ScribeTokens HTR model allocates only 66.9\% of its attention to ink; with pretraining, this rises to 91.2\% (Figure~\ref{fig:attn_htr_sft_scribe}), indicating the model has learned to decode almost entirely from the ink representation.

\paragraph{HTG.} Next-ink-token prediction is itself a generation task: the model must compose strokes into characters and characters into words to predict accurately. Indeed, pretrained models produce recognizable characters and short words, confirming that the objective teaches stroke composition directly. Fine-tuning on HTG then only needs to condition this already-learned generation ability on a text prompt, rather than learning both composition and conditioning from scratch. This is consistent with the large convergence speedups observed in Appendix~\ref{app:speedup}: on IAM, ScribeTokens reaches the baseline loss in a single fine-tuning epoch (\SI{83.0}{\x} speedup).

\paragraph{Point-5.} Unlike token-based models, pretraining can significantly hurt Point-5. Point-5's continuous coordinates already encode spatial structure explicitly, so its embeddings do not suffer from the cold-start problem. Because each prediction step covers only a single coordinate offset, next-ink-vector prediction may be largely solvable by learning momentum alone, without needing to capture higher-level character structure. This shallow objective could bias the model toward local dynamics at the expense of longer-range dependencies: CER rises from 9.43\% to 13.63\% on IAM HTR (Table~\ref{tab:htr}) and from 14.36\% to 26.84\% on DeepWriting HTG (Table~\ref{tab:htg}). Token representations, by contrast, merge multiple steps into each token, so next-token prediction requires genuine compositional understanding that transfers to downstream tasks.

\section{Generated Ink}
\label{app:htg_samples}

\begin{figure}[t]
  \centering
  \includegraphics[width=\textwidth]{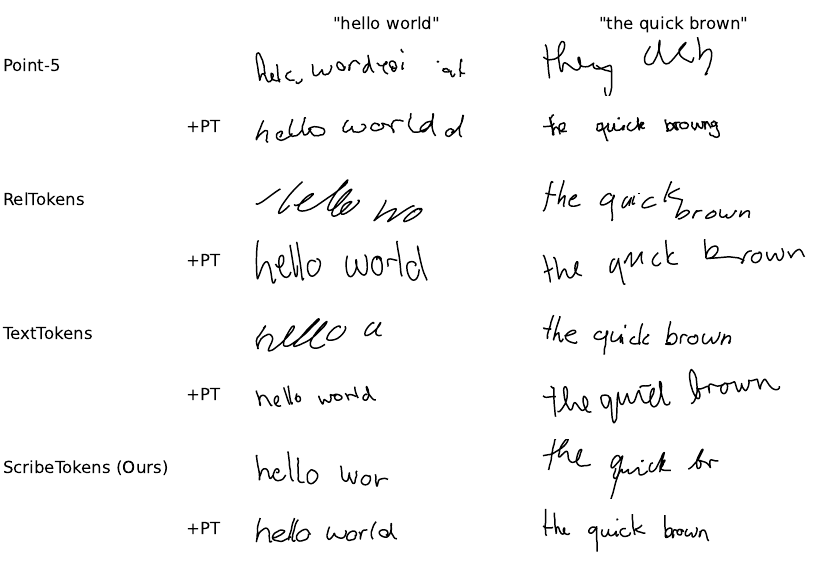}
  \caption{Generated ink samples from all four representations trained on IAM, under autoregressive decoding (temperature 1); all generations are post-processed (Appendix~\ref{app:quantization}). Columns correspond to different text prompts; rows alternate between models trained from scratch and models initialized with next-ink-token prediction pretraining (+PT). Without pretraining, generations are frequently incomplete or illegible; pretraining substantially improves legibility across all methods, with ScribeTokens + PT being the only method to correctly generate both prompts.}
  \label{fig:htg_samples}
\end{figure}

Figure~\ref{fig:htg_samples} shows generated ink from all four representations trained on IAM, under autoregressive decoding (temperature 1); all generations are post-processed (Appendix~\ref{app:quantization}) before visualization. Pretraining improves legibility across the board, but ScribeTokens + PT is the only method to correctly generate both text prompts. These qualitative observations are consistent with the quantitative gains in Table~\ref{tab:htg}.

\paragraph{Failure modes.} Point-5's uncompressed sequences are considerably longer, slowing inference and degrading quality toward the end. RelTokens exhibits collapsed spacing between characters---an artifact of generated \texttt{[UNKNOWN]} tokens that encode no valid displacement. TextTokens generates malformed digit sequences that do not parse into valid coordinate pairs, yielding missing strokes. This is especially common for long strokes such as those in cursive, where the model emits an odd number of coordinate values that cannot be grouped into $(x, y)$ pairs.

\end{document}